%% file: iclr2023_conference.tex
\documentclass{article} 
\usepackage{iclr2023_conference,times}

\input{math_commands.tex}

\usepackage{hyperref}
\usepackage{url}
\usepackage{caption}
\usepackage{subcaption}
 \usepackage{graphicx}
 \usepackage{comment}
 \usepackage{multirow}

\title{Training Large Language Models Efficiently with Sparsity and Dataflow}

\author{Venkat Srinivasan, Darshan Gandhi, Urmish Thakker \& Raghu Prabhakar \\
SambaNova Systems Inc. \\
Palo Alto, CA 94303, USA \\
\texttt{\{venkat.srinivasan\}@sambanova.ai} \\
}

\iclrfinalcopy 
\begin{document}

\maketitle

\begin{abstract}
Large foundation language models have shown their versatility in being able to be adapted to perform a wide variety of downstream tasks, such as text generation, sentiment analysis, semantic search etc. However, training such large foundational models is a non-trivial exercise that requires a significant amount of compute power and expertise from machine learning and systems experts. As models get larger, these demands are only increasing. Sparsity is a promising technique to relieve the compute requirements for training. However, sparsity introduces new challenges in training the sparse model to the same quality as the dense counterparts. Furthermore, sparsity drops the operation intensity and introduces irregular memory access patterns that makes it challenging to efficiently utilize compute resources. This paper demonstrates an end-to-end training flow on a large language model - 13 billion GPT - using sparsity and dataflow. The dataflow execution model and architecture enables efficient on-chip irregular memory accesses as well as native kernel fusion and pipelined parallelism that helps recover device utilization. We show that we can successfully train GPT 13B to the same quality as the dense GPT 13B model, while achieving an end-end speedup of 4.5x over dense A100 baseline. 
\end{abstract}

\section{Introduction}
Foundation models (\cite{rishietal}) in natural language processing (NLP) (e.g. BERT, GPT) and computer vision (e.g. VIT, DALL-E) domain have accelerated deployment of machine learning systems in research and commercial domain. Their key characteristics of self supervision and adaptation, allows a myriad of applications to be built to solve specific problems such as, text generation, sentiment analysis, image segmentation, image detection etc. With the intention of extracting more capabilities out of these models and to train on large corpus of data, researchers have proposed increasing parameters count by orders of magnitude (\cite{DBLP:journals/corr/abs-2005-14165}\cite{wang2022image} \cite{DBLP:journals/corr/abs-1910-10683}). 

Due to power and physical constraints, underlying hardware to train such humongous models does not scale proportional to model parameters (\cite{DBLP:journals/corr/abs-2007-05558, leiserson2020there}), a number of techniques, such as network restructuring (\cite{DBLP:journals/corr/DongHYY17} \cite{DBLP:journals/corr/abs-2004-11886}), network pruning(\cite{DBLP:journals/corr/abs-2003-03033}), network quantization (\cite{NIPS2016_d8330f85}), low rank decomposition \cite{mao-etal-2020-ladabert}, knowledge
distillation \cite{DBLP:journals/corr/abs-1910-01108}, model sparsity (\cite{DBLP:journals/corr/abs-2101-09048} \cite{DBLP:journals/corr/MocanuMSNGL17}) etc. have been explored to handle this computational challenge. Various kinds of sparse techniques have been proposed \cite{DBLP:journals/corr/abs-2103-01636} \cite{DBLP:journals/corr/abs-2001-01969} to reduce computational intensity and to mimic human brain neuron connections (\cite{NIPS1989_6c9882bb},\cite{Azevedo2009}). 

As sparsity techniques continue to evolve and become mainstream in training and inference applications, an entirely new set of challenges are posed to the underlying hardware architecture (\cite{DBLP:journals/corr/abs-2007-00864}). In contrast to coping up to ML computational challenge by mere Tera-FLOPs and memory bandwidth increase, sparse computations demand flexibility, programmability and efficiency from next generation of hardware due to a wide range of possible patterns and training flows (\cite{DBLP:journals/corr/abs-2001-04451}, \cite{DBLP:journals/corr/abs-2112-00029}, \cite{DBLP:journals/corr/HanPNMTECTD16}). A well-balanced system should be able to effectively handle a generally compute intensive dense deployment of a model, a memory intensive highly sparse deployment of a model and variations in between.  A successful deployment of sparse techniques on a friendly architecture can help mitigate current roadblocks, such as immense power, huge machine cost and long training time, in an effective way.

With the expansion of machine learning and artificial intelligent applications and their intrinsic characteristics, a number of computational frameworks have been suggested over time. Some of the examples include Google TPU (\cite{DBLP:journals/corr/JouppiYPPABBBBB17}), Cambricon (\cite{8574528} \cite{7783723}), NVIDIA A100 \cite{nvidia2020a100}, Cerebras CS-2 (\cite{10.1145/3505170.3511036}), Graphcore IPU (\cite{DBLP:journals/corr/abs-1912-03413}) and SambaNova RDU (\cite{sn10-isscc}) in addition to traditional CPU based architectures. While there are a few attempts to evaluate and compare these hardware and software systems (\cite{10024028}, \cite{DBLP:journals/corr/abs-1909-06842}), full scope of their capabilities especially in terms of handling full range of sparse and dense applications, remain unknown. Many of these frameworks also remain proprietary and  unavailable for a generic study in the public domain.

Although lucrative, sparse techniques come with their own set of challenges beyond architectural compatibility. There is a huge spectrum of variables, such as structured \cite{DBLP:journals/corr/WenWWCL16} /semi-structured \cite{DBLP:journals/corr/abs-2102-04010} /unstructured sparsity(\cite{DBLP:journals/corr/HanPTD15}, \cite{DBLP:journals/corr/GuoYC16}), percentages of sparsity \cite{DBLP:journals/corr/abs-2005-07683}, weights/activation sparsity \cite{DBLP:journals/corr/abs-2001-01969} and training schedule \cite{DBLP:journals/corr/HanPNMTECTD16}, that impacts accuracy of a given model compared to a dense only baseline. It is difficult and time-consuming to determine these decision variables to achieve state of the art (SOTA) metrics on a given model. Large language models, such as 13B parameter GPT  are pervasive foundation models in NLP domain, which can support a variety of language applications. In this paper, we use this model to showcase a successful inclusion of sparsity in an end-to-end training flow to achieve comparable accuracy metrics. In the process, we make the following key contributions:

\begin{itemize}
    \item A systematic analysis of the interplay between of sparsity, fusion, and dataflow capabilities.
    \item Evaluation of sparse GPT 13B on SambaNova RDU demonstrating speedups over A100.
    \item Loss, Zero-shot and Few-shot analysis of the sparse 13B GPT model when compared to its dense baseline
\end{itemize}

The rest of this paper is organized as follows. Section~\ref{background} provides a brief background on sparse pretraining and dataflow architectures. Section~\ref{sparsity_dataflow} quantifies the advantages with sparsity and dataflow. Section~\ref{training} describes our methology to train a 13 billion GPT model using sparsity. Section~\ref{evaluation} evaluates the methodology, and section~\ref{conclusion} concludes.

\section{Background and Related Work}
\label{background}

\subsection{Sparse Pretraining}
Sparse training has gained significant interest in recent times. Different sparse training methods have emerged, where sparse weights are maintained during training. Work in this domain includes exploration into various pruning (or growth) criteria such as weight magnitude and sign, random selection and gradient magnitude \cite{neurips2021a, liu2022comprehensive, huang2022dynamic, chen2021pixelated}. This work does not focus on developing a new method for sparse pretraining. Instead, we follow the S2D methodology developed in \cite{chen2021pixelated}. This work differentiates from these previous work in terms of the scale of the model being explored (13 billion parameters), the hardware used for the exploration (dataflow architecture) and its implication. All prior work has either focused on training on more traditional hardware - TPUs, GPUs or CPUs, which are significantly different than a more efficient and faster dataflow hardware.

\subsection{Dataflow Architectures}
Hardware dataflow accelerator architectures have recently emerged as a promising design choice to keep up with the ever-increasing compute demands of large language models. Dataflow accelerators are typically composed of programmable compute and memory units placed in a programmable interconnect fabric. Unlike a conventional architecture that executes programs as instruction streams with a global program counter, dataflow architecture components are often statically configured. Dataflow architectures avoid the power and area overheads of traditional instruction management, and enable automatic kernel fusion with pipelining without manually writing fused kernels. Input compute graphs are lowered by dataflow compilers into a graph of primitive compute and memory units, which then gets placed and routed on the available physical units on the target dataflow hardware. Numerous types of dataflow architectures have been proposed both in industry and academia in the past decade~\cite{plasticine, cgra-survey}. The architectures explore different design points in the granularity of compute, on-chip memory system, and flexibility in the interconnect. 

\begin{figure}
     \centering
     \includegraphics[scale=0.45]{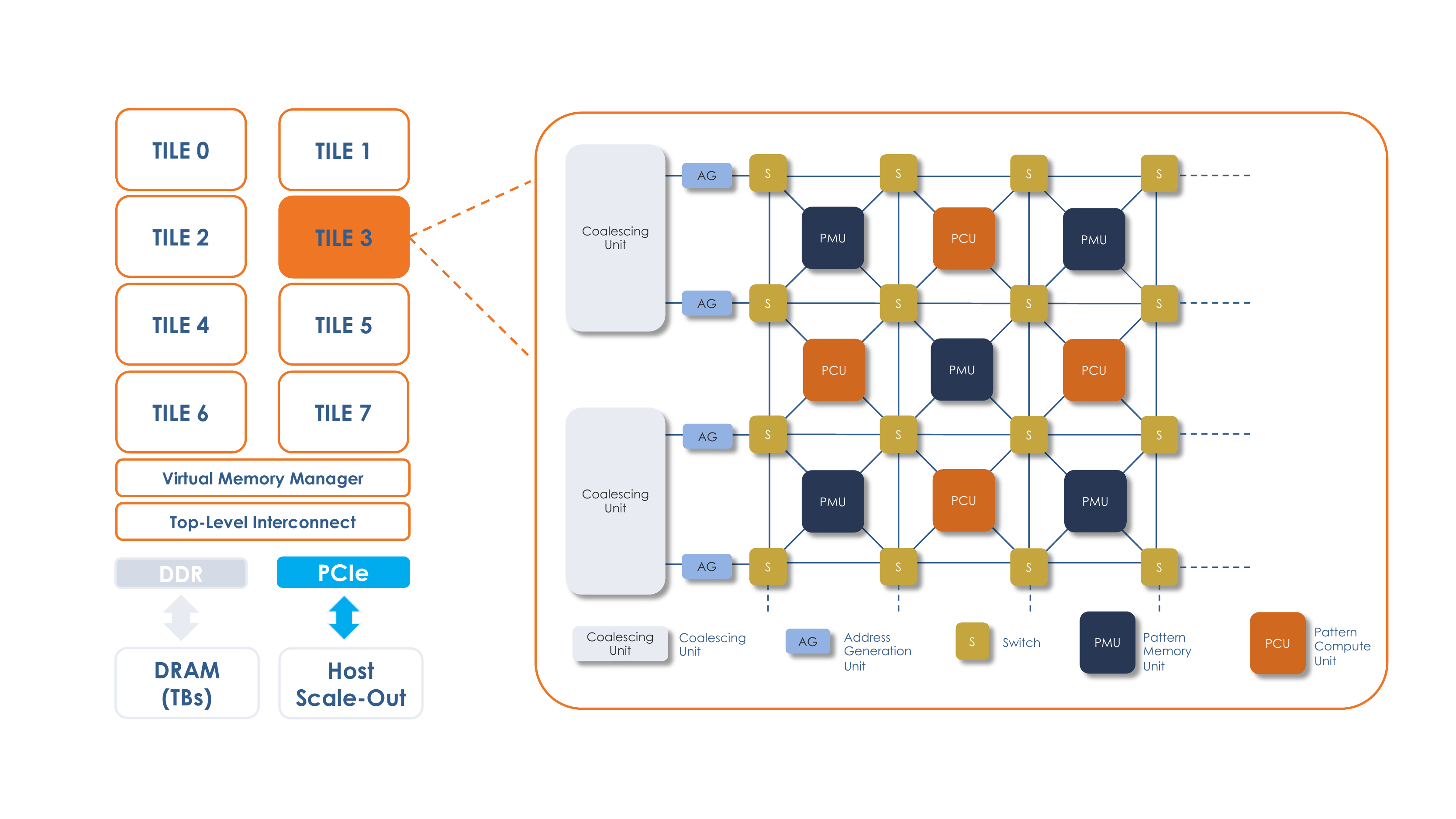}
     \caption{RDU Dataflow Architecture. PCUs are the programmable compute elements, PMUs form the on-chip memory system, and S represents the programmable interconnect. DDR and other IO is accessed via AGs and CUs.}
     \label{fig:rdu}
\end{figure}

 In this paper, we study the impact of sparsity in the context of SambaNova's Reconfigurable Dataflow Unit (RDU)~\cite{sn10-hc, sn10-isscc}. Figure~\ref{fig:rdu} describes the high-level architecture of the SambaNova RDU. The RDU is organized as multiple tiles of compute, memory, and interconnect components. Within each tile, PCUs (orange) are the programmable compute elements that contain multiple pipeline stages of SIMD ALUs. PMUs (blue) are distributed software-managed on-chip scratchpad memories with programmable address generation and tensor transformations. PCUs and PMUs are connected to each other to form a software pipeline via the programmable interconnect switches S (yellow). DDR and other IO is accessed via AGs and CUs (gray).

\section{Sparsity and Dataflow}
\label{sparsity_dataflow}
\begin{figure}
     \centering
     \includegraphics[scale=0.55]{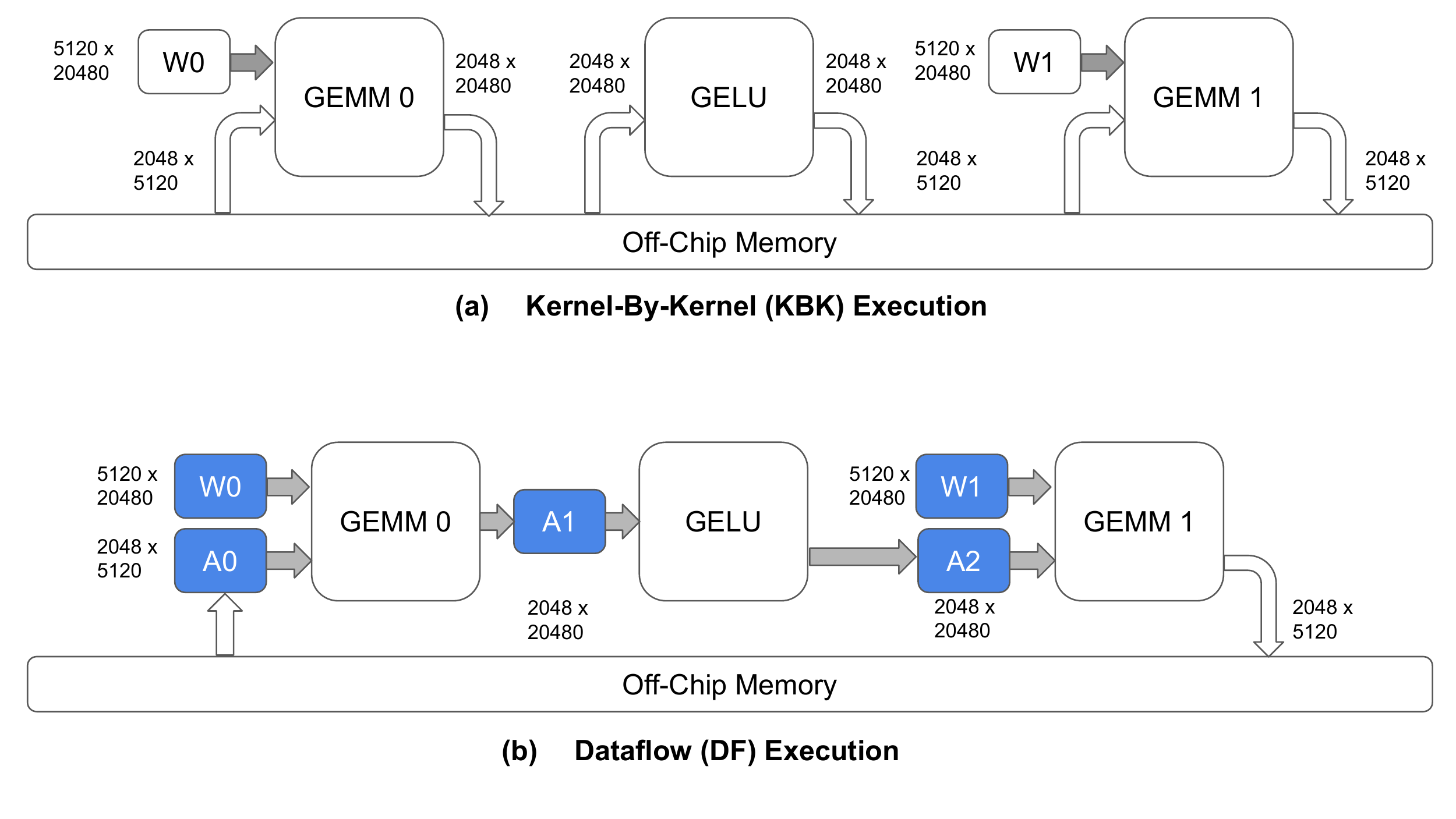}
     \caption{Kernel-By-Kernel (KBK) vs. Dataflow (DF) execution for a simple example with GEMM followed by GELU followed by GEMM. White arrows represent traffic to off-chip memory, gray arrows represent on-chip traffic. Blue boxes represent on-chip memory, white boxes represent on-chip compute resources. We use matrix dimensions from GPT 13B in this example. Each edge / box is labeled with the Tensor dimensions being accessed.}
     \label{fig:kbk_vs_df}
\end{figure}

In this section, we describe and quantify two execution models: \emph{Kernel-By-Kernel} (KBK) and \emph{Dataflow} (DF) with respect to weight sparsity. In KBK execution, operators in a compute graph are executed one at a time. Intermediate results between two operators are exchanged through off-chip memory. The compute and memory resources available can be used to execute each operation fully in parallel. In DF execution, multiple operators are connected together to form a dataflow pipeline. Intermediate results between operators are exchanged through on-chip double buffers. Consequently, the compute and memory resources available need to be shared between all the operators in the pipeline. 

Figure~\ref{fig:kbk_vs_df} pictorially shows a simple compute graph involving the FFN block in GPT encoder. This constitutes two GEMM computations and a GELU operation between them. We use tensor dimensions from a 13 billion GPT encoder as an example for the remainder of this section. Figure~\ref{fig:kbk_vs_df}(a) shows KBK while figure~\ref{fig:kbk_vs_df}(b) shows DF execution. Arrows show both off-chip (white) and on-chip (gray) traffic. The edges are labeled with the Tensor dimensions being transferred. Blue boxes show Tensors in on-chip memory in DF execution.

Table~\ref{kbk-vs-df} quantifies the impact of weight sparsity on KBK vs DF.
We study three parameters: off-chip memory bandwidth (BW), on-chip memory capacity (M), and achievable speedup (X) for various sparsity levels (S). 
For both KBK and DF, we assume a compute capability of 300 TFLOPs.
\emph{BW} shows the minimum bandwidth required to fully utilize the available compute capabilities (100\% of TFLOPS). While we recognizing that 100\% utilization is impractical in general, this analysis enables identifying and quantifying fundamental bottlenecks that are unavoidable even with perfect execution of sparse GEMMs. \emph{M} shows the total memory capacity required by DF to construct the dataflow pipeline. \emph{X} shows the achievable speedup when latencies of all operations are factored in.

From Table~\ref{kbk-vs-df}, we can see that \emph{BW} requirement for KBK increases proportional to sparsity. Furthermore, the bandwidth requirements for KBK can be an order of magnitude higher than DF. This observation has important system-level implications: an accelerator architecture built around KBK execution requires a significantly higher memory bandwidth than DF to fully exploit sparsity. The high BW requirement stems from the fact sparsity is reducing total compute which makes GEMMs execute faster, and that all intermediate results are stored and loaded back from off-chip memory, resulting in more memory traffic. The DF model is able to successfully capture data locality between operations on-chip, and hence does not incur the same penalty. DF execution fully utilizes compute TFLOPS by exploiting pipeline parallelism between operators. DF gets the effect of kernel fusion natively by enabling the construction of such operator pipelines in software. The impact of operator fusion have been studied in previous literature too~\cite{asplos22-google}. Low BW requirements for DF enable such accelerators to be built with dense, lower bandwidth off-chip memory technologies like DDR with TBs of capacity. The larger capacity enables more efficient mapping of large foundational models without complex sharding requirements.

Table~\ref{kbk-vs-df} also shows that DF needs comparatively larger on-chip memory capacity. Note that this example effectively serves as the upper bound for \emph{M}, as this study excludes commonly used techniques like tiling to reduce the on-chip memory requirement on DF. The blue boxes in Figure~\ref{fig:kbk_vs_df} shows that the on-chip memory is being used to hold weights and intermediate results between pipeline stages. Furthermore, these buffers need to be double-buffered to decouple the producing stage from the consuming stage. As sparsity reduces the size of weights, \emph{M} goes down as sparsity increases.

Finally, Table~\ref{kbk-vs-df} also compares achievable speedups on KBK vs. DF. The speedup calculation assumes that KBK has a peak off-chip bandwidth of 2 TB/s, and that both KBK and DF can run sparse GEMMs at full efficiency. As sparsity increases, sparse GEMMs get proportionally faster. However, note that the output of each sparse GEMM is still a dense tensor. Due to Amdahl's law, total speedup gets impacted by the memory-bound GELU which is not helped by sparsity. Even if GELU is able to fully utilize all 2 TB/s bandwidth on KBK, we see that the total speedup scales to only 12.9x on KBK. In contrast, DF speedup can scale almost linearly with sparsity. This is because, while DF loses some TFLOPs to compute GELU, it is able to proportionally parallelize GELU to match the sparse GEMM throughput. This is enabled by larger on-chip bandwidth, a flexible compute and memory system that allows processing multiple streams of data from A1 in Figure~\ref{fig:kbk_vs_df}(b). 

In summary, the benefits of sparsity at the compute graph level can vary widely between KBK and DF. KBK is more sensitive to available off-chip bandwidth and other memory-bound operations in the model like GELU. DF requires large memory capacity, but can sustain higher overall utilization with an order of magnitude lesser off-chip bandwidth. The next section describes how these insights are utilized to train a 13 billion GPT model on the SambaNova Reconfigurable Dataflow Unit (RDU).

\begin{table}[t]
\caption{Kernel-By-Kernel (KBK) and Dataflow (DF) execution impact for a sequence of GEMM - GELU - GEMM operations. The table shows off-chip memory bandwidth and on-chip memory capacity requirements for various weight sparsity levels.}
\label{kbk-vs-df}
\begin{center}
\begin{tabular}{r|rr|r|rrr}
\multicolumn{1}{c|}{\bf S: Sparsity }  &\multicolumn{2}{c|}{\bf BW: Bandwidth (GB/s)} & \multicolumn{1}{c|}{\bf M: On-Chip Memory (MB)} & \multicolumn{3}{c}{\bf X: Speedup vs. Dense} \\
                      & KBK     & DF                &  DF    & Ideal     & KBK   & DF    \\
\hline \\
(Dense) 0\%           & 134     & 17                & 839   & 1         & 1     & 1     \\
50\%                  & 266     & 32                & 629   & 2         & 1.94  & 2  \\
87.5\%                & 1057    & 119               & 472   & 8         & 6.65  & 7.9  \\
95\%                  & 2639    & 295               & 440   & 20        & 12.9  & 19.6  \\

\end{tabular}
\end{center}
\end{table}

\begin{table}[t]
\begin{tabular}{|l|l|l|l|l|}
\hline
\textbf{Benchmark}                          & \textbf{Task Name} & \textbf{Metric} & \textbf{Dense 13B} & \textbf{S2D 13B} \\ \hline
\multirow{4}{*}{Cloze/Completion}           & LAMBADA            & Perplexity      & 6.47        & 6.13           \\ \cline{2-5} 
                                            & LAMBADA            & Accuracy        & 62.12\%            & 61.85\%          \\ \cline{2-5} 
                                            & HellaSwag          & Accuracy        & 43.96\%            & 44.50\%          \\ \cline{2-5} 
                                            & HellaSwag          & Accuracy (Norm) & 57.28\%            & 58.28\%          \\ \hline
\multirow{9}{*}{Common Sense Reasoning}     & TriviaQA           & Accuracy        & 7.81\%             & 8.69\%           \\ \cline{2-5} 
                                            & OpenBookQA         & Accuracy        & 22.80\%            & 23.60\%          \\ \cline{2-5} 
                                            & OpenBookQA         & Accuracy (Norm) & 35.00\%            & 36.40\%          \\ \cline{2-5} 
                                            & ARC (Easy)         & Accuracy        & 59.81\%            & 60.56\%          \\ \cline{2-5} 
                                            & ARC (Easy)         & Accuracy (Norm) & 52.44\%            & 55.93\%          \\ \cline{2-5} 
                                            & ARC (Challenge)    & Accuracy        & 27.30\%            & 28.41\%          \\ \cline{2-5} 
                                            & ARC (Challenge)    & Accuracy (Norm) & 29.44\%            & 31.32\%          \\ \cline{2-5} 
                                            & PiQA               & Accuracy        & 72.58\%            & 72.96\%          \\ \cline{2-5} 
                                            & PiQA               & Accuracy (Norm) & 74.32\%            & 73.12\%          \\ \hline
\multirow{4}{*}{Natural Language Inference} & ANLI\_R1           & Accuracy        & 32.70\%            & 36.10\%          \\ \cline{2-5} 
                                            & RTE                & Accuracy        & 55.60\%            & 53.07\%          \\ \cline{2-5} 
                                            & MNLI-matched       & Accuracy        & 35.18\%            & 33.09\%          \\ \cline{2-5} 
                                            & MNLI-mismatched    & Accuracy        & 35.60\%            & 33.97\%          \\ \hline
Winograd-Style Tasks                        & Winogrande         & Accuracy        & 59.59\%            & 58.96\%          \\ \hline
\end{tabular}
\caption{Zero-shot accuracy on various benchmarks}
\label{tab:zeroshot}
\end{table}

\begin{table}[]
\begin{tabular}{|l|l|l|l|l|}
\hline
\textbf{Benchmark}                          & \textbf{Task Name} & \textbf{Metric} & \textbf{Dense 13B} & \textbf{S2D 13B} \\ \hline
\multirow{3}{*}{Cloze/Completion}           & LAMBADA            & Accuracy        & 54.24\%            & 54.45\%          \\ \cline{2-5} 
                                            & HellaSwag          & Accuracy        & 43.79\%            & 44.70\%          \\ \cline{2-5} 
                                            & HellaSwag          & Accuracy (Norm) & 57.39\%            & 58.99\%          \\ \hline
\multirow{9}{*}{Common Sense Reasoning}     & TriviaQA           & Accuracy        & 13.70\%            & 16.11\%          \\ \cline{2-5} 
                                            & OpenBookQA         & Accuracy        & 23.20\%            & 26.00\%          \\ \cline{2-5} 
                                            & OpenBookQA         & Accuracy (Norm) & 34.60\%            & 36.20\%          \\ \cline{2-5} 
                                            & ARC (Easy)         & Accuracy        & 62.16\%            & 64.86\%          \\ \cline{2-5} 
                                            & ARC (Easy)         & Accuracy (Norm) & 61.11\%            & 63.89\%          \\ \cline{2-5} 
                                            & ARC (Challenge)    & Accuracy        & 30.29\%            & 30.38\%          \\ \cline{2-5} 
                                            & ARC (Challenge)    & Accuracy (Norm) & 33.02\%            & 34.47\%          \\ \cline{2-5} 
                                            & PiQA               & Accuracy        & 72.47\%            & 73.39\%          \\ \cline{2-5} 
                                            & PiQA               & Accuracy (Norm) & 73.78\%            & 74.65\%          \\ \hline
\multirow{4}{*}{Natural Language Inference} & ANLI\_R1           & Accuracy        & 30.60\%            & 32.90\%          \\ \cline{2-5} 
                                            & RTE                & Accuracy        & 49.46\%            & 45.85\%          \\ \cline{2-5} 
                                            & MNLI-matched       & Accuracy        & 34.98\%            & 31.10\%          \\ \cline{2-5} 
                                            & MNLI-mismatched    & Accuracy        & 36.45\%            & 31.27\%          \\ \hline
Winograd-Style Tasks                        & Winogrande         & Accuracy        & 60.85\%            & 60.69\%          \\ \hline
\end{tabular}
\caption{4 Shot accuracy on various benchmarks}
\label{tab:4shot}
\end{table}

\section{Training Methodology}
\label{training}
We train a 13 billion GPT model from scratch on the C4 dataset. We use the recipe from \cite{NEURIPS2020_1457c0d6} to train the model. We train the dense version of the model using a batch size of 1024, sequence length of 2048 and train for 150,000 steps. We use a learning rate of 3e-5 with a warm-up of 3000 steps. We use a similar recipe for our sparse version of the model. We train the sparse model for some number of the steps and "densify" and train the model using the dense version for the rest of the steps (S2D) \cite{chen2021pixelated}. We vary the sparse steps and chose the one that leads to iso-accuracy. During the sparse phase, we experiment with sparsity levels of 75-87.5\% sparsity. We train the model end-end on the SambaNova hardware. 

\section{Evaluation}
\label{evaluation}

\begin{figure}
     \centering
     \includegraphics[scale=0.4]{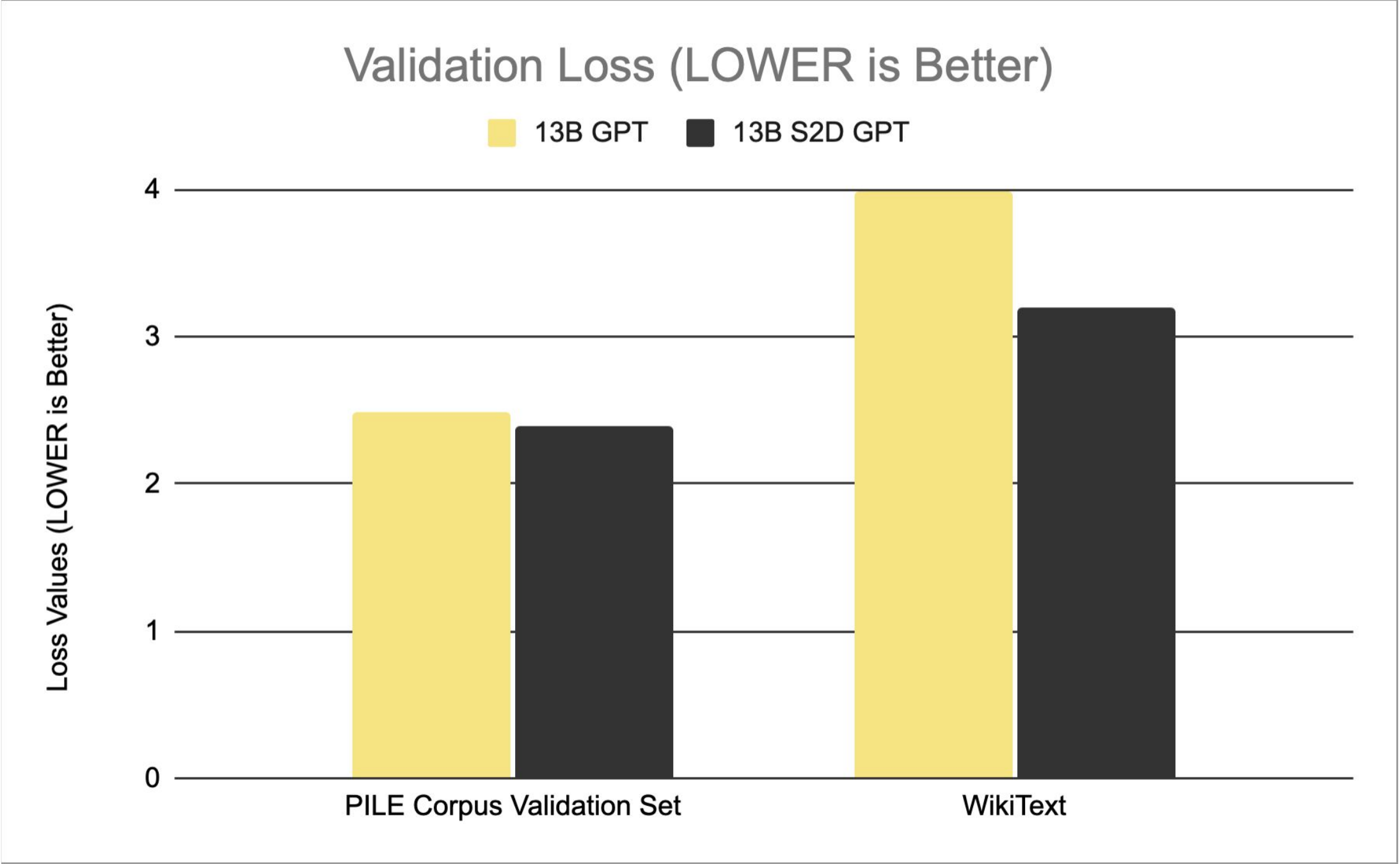}
     \caption{Loss on a held out dataset}
     \label{fig:c4loss}
\end{figure}
We evaluate the dense and S2D version of the 13 billion GPT model by comparing the loss on a held out evaluation C4 dataset and zero shot accuracy on LAMBADA \cite{lambada}, HellaSwag \cite{hellaswag}, TriviaQA \cite{JoshiTriviaQA2017}, OpenBookQA \cite{OpenBookQA2018}, PiQA \cite{Bisk2020}, RTE \cite{wang2019glue}, Winogrande \cite{ai2:winogrande}, COPA \cite{gordon-etal-2012-semeval} and ANLI\_R1 \cite{williams-etal-2020-anlizing}. Results for the same can be viewed in figure~\ref{fig:c4loss} and tables ~\ref{tab:zeroshot} and ~\ref{tab:4shot}. For both zero shot accuracy and loss on a held out dataset, the S2D version achieves the same accuracy as the dense version. Overall, our pipeline on RDU HW gives us a speed-up of 4.5x over A100.

\section{Conclusion}
\label{conclusion}
In this paper, we describe a method to train large language models efficiently with sparsity and dataflow. We describe the locality and pipelined parallelism advantages with dataflow execution over kernel-by-kernel execution with various sparsity levels, and quantify the performance benefits achievable with dataflow execution. We then describe a flow using S2D to train a 13 billion GPT model from scratch on the C4 dataset, and evaluate on a variety of downstream tasks on the SambaNova RDU. We show that our S2D version achieves the same accuracy as the dense model while achieving an end-to-end speedup of 4.5x over the dense model on A100

\bibliography{iclr2023_conference}
\bibliographystyle{iclr2023_conference}

\end{document}

%% file: math_commands.tex
\usepackage{amsmath,amsfonts,bm}

\def\eqref#1{equation~\ref{#1}}

\def\1{\bm{1}}

\DeclareMathAlphabet{\mathsfit}{\encodingdefault}{\sfdefault}{m}{sl}
\SetMathAlphabet{\mathsfit}{bold}{\encodingdefault}{\sfdefault}{bx}{n}

